\documentclass[letterpaper, 10 pt, onecolumn]{ieeeconf}

\IEEEoverridecommandlockouts
\usepackage{amsmath}
\usepackage{amssymb}
\usepackage{booktabs}
\usepackage{multirow}
\usepackage{xcolor}
\usepackage{graphicx}
\usepackage{float}

\title{\LARGE \bf
Smart Fusion for Adaptive Visual-IMU Weighting\\
in cuVSLAM Sparse Bundle Adjustment
}

\author{Wonju Lee$^{1}$, Dara Molloy$^{1}$, Abhilash Nand Kumar$^{1}$, Onias Catelo$^{1}$\\
Petar Trslic$^{1}$, David Adley$^{1}$, Conor Ferguson$^{1}$, and Hesam Rabeti$^{2}$%
\\
{\footnotesize
$^{1}$Analog Devices Inc., \{wonju.lee,dara.molloy,abhilash.nandkumar,onias.castelo,petar.trslic,david.adley,conor.ferguson\}@analog.com \\
$^{2}$NVIDIA, hrabeti@nvidia.com
}
}

\begin{document}

\maketitle
\thispagestyle{empty}
\pagestyle{empty}

\begin{abstract}
In cuVSLAM, visual and inertial measurement unit (IMU) residuals are fused in sparse bundle adjustment (SBA) with a weighting fixed by pre-calibrated noise covariances, leaving no mechanism to adapt the visual-inertial balance at runtime.
We introduce a scalar $\alpha\geq1$ that upscales the visual information matrix relative to the IMU preintegration constraint. Across the candidate set $\{1.0,2.0,5.0,10.0,20.0\}$, the resulting trajectories form two empirical performance regimes. The \textsc{imu-trust} regime contains $\alpha\in\{1,2\}$ near the calibrated baseline, while the \textsc{vision-trust} regime contains $\alpha\in\{10,20\}$. The intermediate value $\alpha=5$ is treated as a transition case. Differences within each regime are small relative to those observed between the regimes.
This structure motivates reducing $\alpha$-selection to a binary decision.
We train a lightweight Visual Quality Estimator (VQE), implemented as a multilayer perceptron with 914 trainable parameters. It maps a 10-dimensional visual quality feature vector assembled from statistics already maintained by the tracker to a per-SBA regime prediction.
The proposed method reduces mean RMSE APE by $27.7\%$ and by $14.9\%$ under leave-one-out cross-validation (LOOCV) on EuRoC MAV and TUM-VI Room, respectively.
\end{abstract}

\begin{figure}[H]
  \centering
  \includegraphics[width=0.8\columnwidth]{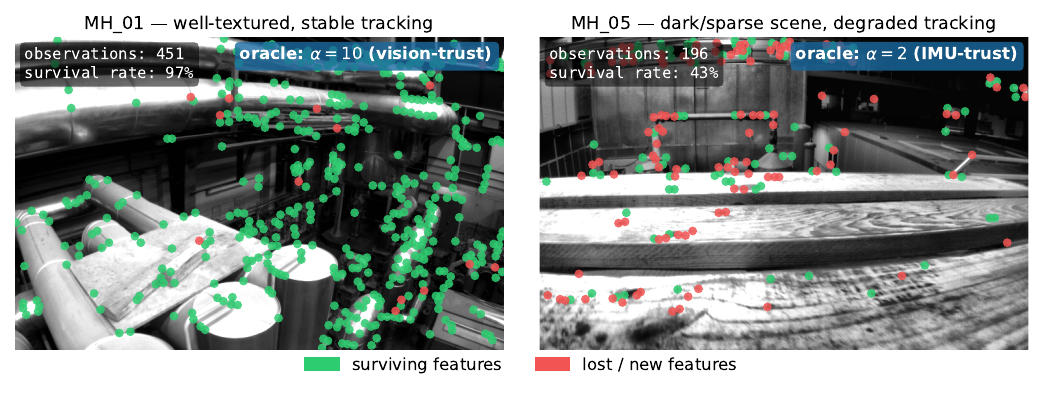}
  \vspace{-4mm}
  \caption{Feature quality contrast on EuRoC Machine Hall (MH\_01 vs.\ MH\_05).
    Corners are detected with GFTT on the raw image.
    The fraction of green points is chosen to match the feature survival rate recorded in the SBA log under the greedy-reference assignment.
    The coloring represents a per-frame aggregate and should not be interpreted as a per-point tracking outcome.
    The greedy reference assigns $\alpha\!=\!10$ (\textsc{vision-trust}) to the MH\_01 segment and $\alpha\!=\!2$ (\textsc{imu-trust}) to the MH\_05 segment.}
  \label{fig:feature_quality}
  \vspace{-2mm}
\end{figure}

\section{INTRODUCTION}
Tightly-coupled visual-inertial odometry (VIO) fuses reprojection residuals and IMU preintegration constraints in a joint optimization~\cite{c_cuvslam,c_vins,c_orbslam3}.
The relative weight assigned to the two residual types affects the influence of each sensor modality on the optimized trajectory.
Increasing the visual weight under degraded observations can increase the effect of visual measurement errors. Conversely, a low visual weight may underuse geometric constraints when the observations are reliable.
An appropriate balance can depend on scene and motion conditions, yet most systems determine it during calibration and retain it during operation.

In the original cuVSLAM~\cite{c_cuvslam} formulation, visual and IMU residuals are combined under a fixed weighting derived from pre-calibrated noise covariances (corresponding to $\alpha\!=\!1.0$), with no mechanism to adjust this balance at runtime.
This fixed weighting is appropriate when the calibrated noise model remains representative of the measurements. It may be less appropriate when scene conditions such as sparse texture or large depth variation increase reprojection error, or when motion blur and illumination changes temporarily degrade tracking.
Under these conditions, retaining the visual weight near its nominal calibrated level can limit the influence of degraded measurements, but the original formulation does not perform this adaptation.
Fig.~\ref{fig:feature_quality} illustrates contrasting visual-quality conditions and the corresponding regimes selected by the greedy reference.

We present Smart Fusion, which introduces a global fusion weight $\alpha$ to scale the visual information matrix relative to the IMU constraint.
Smart Fusion uses a lightweight Visual Quality Estimator (VQE), implemented as an MLP, to map tracker-derived visual quality features to a regime prediction and set $\alpha$ before each SBA call.
The design is motivated by the empirical grouping observed over $\mathcal{A}=\{1.0, 2.0, 5.0, 10.0, 20.0\}$. The \textsc{imu-trust} regime contains $\alpha\in\{1,2\}$ and the \textsc{vision-trust} regime contains $\alpha\in\{10,20\}$. On many evaluated sequences, within-regime differences are small relative to between-regime differences.
On EuRoC MH\_01, the APE difference between the two regimes is $47\%$, while the tested values within each regime produce similar APE. This observation motivates the binary formulation used in Smart Fusion (Fig.~\ref{fig:basin}).
The VQE operates on a 10-dimensional visual quality feature vector assembled from statistics already maintained by the tracker.
Training labels are generated by a sequential non-causal greedy reference that evaluates final full-trajectory APE for each sequence segment.

We make two contributions.
First, we characterize the two empirical performance regimes observed when varying $\alpha$ in cuVSLAM SBA and formulate their selection as binary classification.
Second, we present Smart Fusion, a 914-parameter MLP that predicts a regime at each SBA call from tracker-internal visual quality features. The model is trained on sequential greedy-reference labels derived from trajectory APE.
Smart Fusion reduces mean RMSE APE by $27.7\%$ on EuRoC MAV and by $14.9\%$ on TUM-VI Room, both under leave-one-out cross-validation.

\section{RELATED WORK}

\paragraph{Optimization-based VIO}
Tightly-coupled sliding-window optimizers such as VINS-Mono~\cite{c_vins}, ORB-SLAM3~\cite{c_orbslam3}, OKVIS~\cite{c_okvis}, and cuVSLAM~\cite{c_cuvslam} minimize a joint cost over reprojection and IMU preintegration residuals.
The relative weighting between visual and inertial terms is typically determined by pre-calibrated noise covariances and held fixed during operation.

\paragraph{Noise model and weight adaptation}
Several works adjust sensor noise models online.
Kelly and Sukhatme~\cite{c_imu_noise} perform online visual-inertial localization, mapping, and sensor-to-sensor extrinsic self-calibration, and Cong et al.~\cite{c_cov_adapt} adapt per-feature visual covariances for UAV VIO.
VINS-Mah~\cite{c_vinsmah} and Sibley et al.~\cite{c_robust} propose sliding-window estimators with robust cost functions to down-weight individual high-residual observations.
These approaches operate at the per-feature or per-residual level. Smart Fusion instead adjusts the global visual-to-IMU ratio at each SBA call and therefore provides a complementary form of adaptation.

\paragraph{Learned components in VIO}
Nascivera et al.~\cite{c_nascivera} adapt front-end hyperparameters via an image-conditioned learned policy, and AdaGate~\cite{c_adagate} uses an oracle-supervised MLP to adapt MSCKF feature-gating thresholds from filter state.
Smart Fusion follows a similar oracle-supervised approach but operates on the optimization-based backend and uses a binary representation of the observed $\alpha$ regimes.
Methods such as DROID-SLAM~\cite{c_droidslam} incorporate learned recurrent update operators into dense bundle adjustment. This differs in scope from the lightweight backend parameter adaptation considered here.


\begin{figure}[t]
  \centering
  \includegraphics[width=0.65\columnwidth]{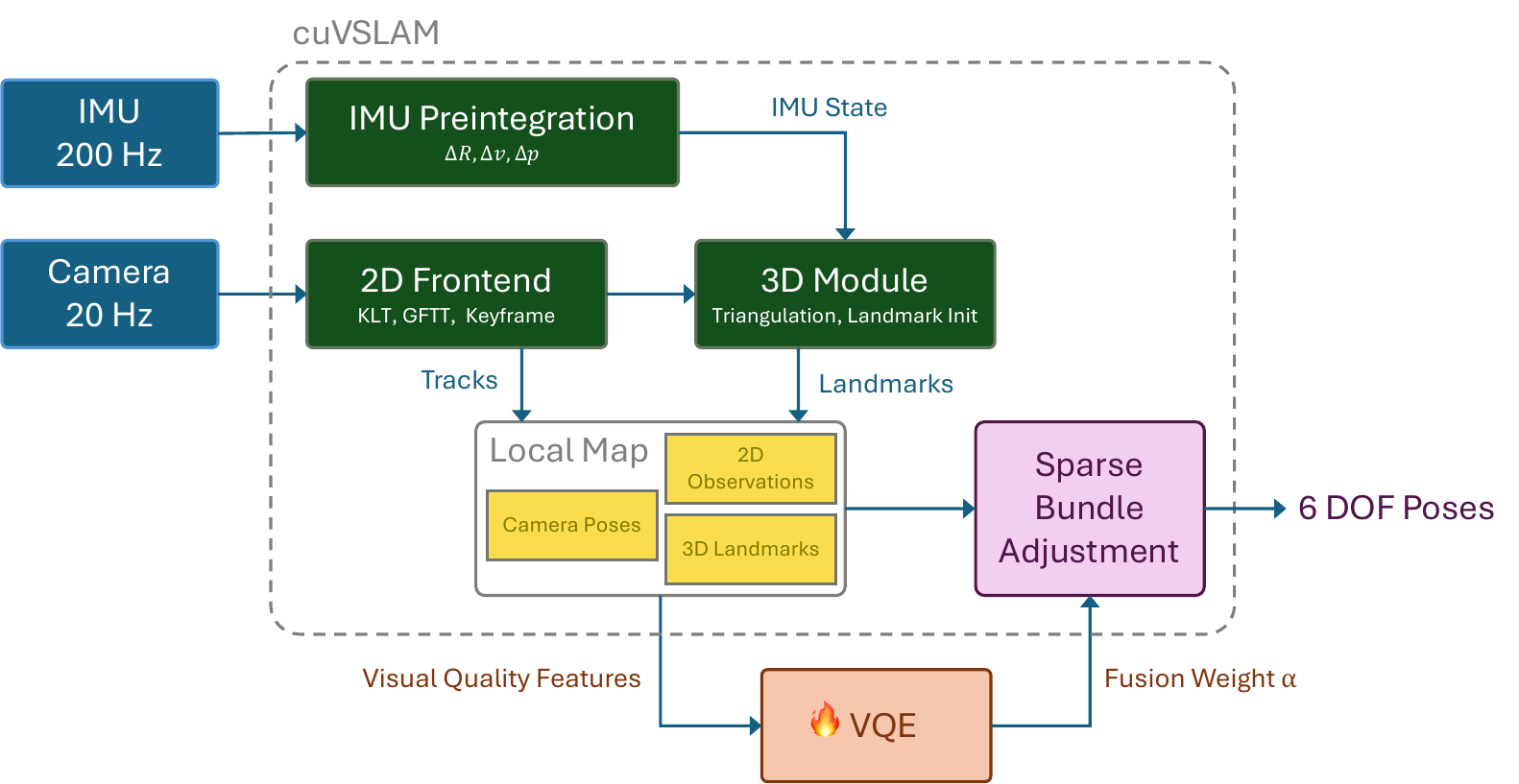}
  \caption{cuVSLAM front-end pipeline with Smart Fusion.
    The 2D Frontend tracks features and selects keyframes.
    The 3D Module initializes landmarks via stereo triangulation using both 2D tracks and IMU state.
    SBA runs asynchronously minimizing the joint visual-inertial cost of~\eqref{eq:sba}.
    The Visual Quality Estimator (VQE) within Smart Fusion predicts $\alpha$ from 10-dimensional visual quality features before each SBA call, replacing the fixed default $\alpha\!=\!1.0$.}
  \label{fig:arch}
  \vspace{-2mm}
\end{figure}

\section{Smart Fusion}

Fig.~\ref{fig:arch} shows the cuVSLAM front-end and the point at which Smart Fusion is applied.
SBA is triggered asynchronously at each new keyframe.
Immediately before each optimization call, Smart Fusion assembles a visual quality feature vector from tracker-internal statistics and queries the VQE to set $\alpha$ for that call.
The cuVSLAM solver structure is unchanged.
Smart Fusion adds a scalar multiplier to the visual information matrix and one forward pass through a 914-parameter network.

\subsection{Alpha in cuVSLAM Bundle Adjustment}

cuVSLAM minimizes the following cost over a sliding window of states and landmarks.
\begin{equation}
  (\hat{\mathbf{S}}_{0:K},\{\hat{\mathbf{p}}_j\}) =
  \arg\min_{\mathbf{S}_{0:K},\{\mathbf{p}_j\}} \left[
    \sum_{i} \bigl\|\mathbf{r}^{\mathrm{imu}}_{i}\bigr\|^{2}_{\boldsymbol{\Omega}_{\mathrm{IMU}}}
    + \alpha \sum_{i,j,k} \bigl\|\mathbf{r}^{\mathrm{repr}}_{ijk}\bigr\|^{2}_{\boldsymbol{\Omega}_{\mathrm{vis}}}
  \right]
  \label{eq:sba}
\end{equation}
where $\|\mathbf{r}\|^2_{\boldsymbol{\Omega}}=\mathbf{r}^\top\boldsymbol{\Omega}\mathbf{r}$ is the squared Mahalanobis norm,
$\mathbf{S}_{0:K}=\{\mathbf{T}^{wb}_i,\mathbf{v}_i,\mathbf{b}_i\}$ is the sliding-window state (world-from-body poses, velocities, IMU biases),
$\mathbf{p}_j\in\mathbb{R}^3$ is the world position of landmark $j$,
$\mathbf{r}^{\mathrm{imu}}_{i}$ is the IMU preintegration residual between consecutive states~\cite{c_preint} with information matrix $\boldsymbol{\Omega}_{\mathrm{IMU}}$,
$\mathbf{r}^{\mathrm{repr}}_{ijk}=\pi(\mathbf{T}^{c_k b}\mathbf{T}^{bw}_{i}\mathbf{p}_{j})-\mathbf{o}_{ijk}$ is the reprojection residual of landmark $j$ in camera $k$ at keyframe $i$ ($\mathbf{T}^{bw}_{i}\!\in\!SE(3)$ body-from-world pose, $\mathbf{T}^{c_k b}$ fixed camera-$k$-from-body extrinsic, $\mathbf{o}_{ijk}$ observed keypoint),
and $\boldsymbol{\Omega}_{\mathrm{vis}}=\sigma^{-2}\mathbf{I}$ is the fixed pixel-noise precision ($\sigma\!\approx\!1\,\text{px}$).

The IMU term aggregates raw measurements between keyframes into a $(\Delta\mathbf{R}, \Delta\mathbf{v}, \Delta\mathbf{p})$ preintegrated constraint~\cite{c_preint} with a $9\!\times\!9$ information matrix.
$\alpha\!=\!1.0$ weights the two terms according to their nominal noise covariances.
Increasing $\alpha$ above $1.0$ raises the relative influence of visual observations.
The candidate set explored in this work is $\mathcal{A}=\{1.0, 2.0, 5.0, 10.0, 20.0\}$, which restricts $\alpha$ to values at or above the nominal setting and therefore never down-weights the visual term relative to the baseline.
Values below $\alpha\!=\!1.0$ were not explored in this work.
The greedy reference produces similar full-trajectory APE for $\alpha\!=\!1.0$ and $\alpha\!=\!2.0$ across the evaluated sequences. We therefore treat these values as members of the same performance regime.
The behavior of values below $\alpha\!=\!1.0$ is not examined in this work.

The preferred regime depends on the quality of the visual observations.
When the features are abundant, well distributed, and stable, the \textsc{vision-trust} regime assigns greater influence to the available geometric constraints.
When visual quality degrades through sparse observations, high reprojection error, or low feature survival, a large $\alpha$ can increase the effect of visual measurement errors and shift the optimized trajectory away from the IMU-predicted state.
Under such conditions, the \textsc{imu-trust} regime may be preferable.

\begin{figure}[t]
  \centering
  \includegraphics[width=0.6\columnwidth]{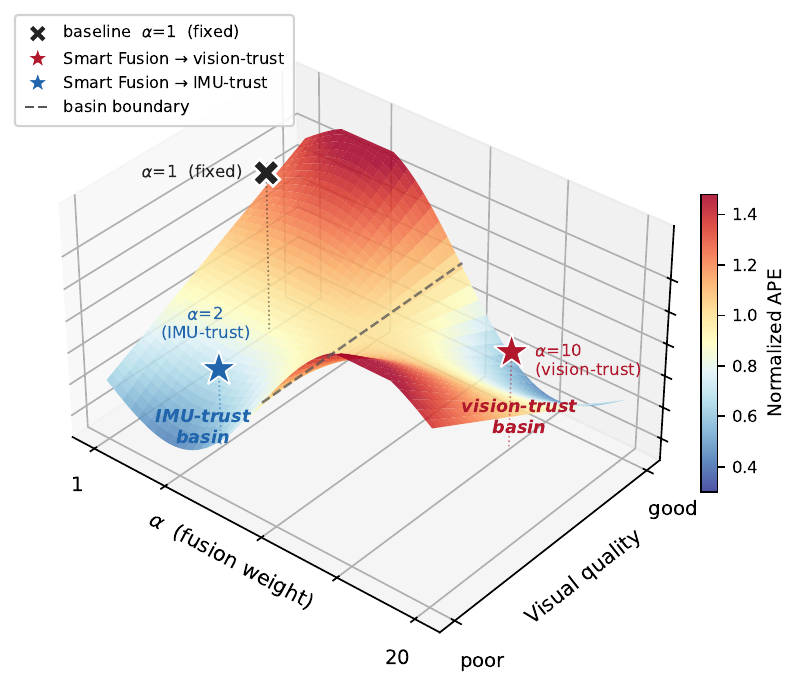}
  \vspace{-3mm}
  \caption{Conceptual APE landscape as a function of $\alpha$ and visual quality.
    The \textsc{imu-trust} regime at low $\alpha$ and the \textsc{vision-trust} regime at high $\alpha$ are separated by a transition region near $\alpha\!=\!5$.
    Under good-quality conditions the fixed baseline $\alpha\!=\!1.0$ ($\times$) sits in the \textsc{imu-trust} regime while the greedy-reference selection lies in the \textsc{vision-trust} regime.
    Smart Fusion aims to select the appropriate regime ($\bigstar$) based on the observed visual quality.}
  \label{fig:basin}
  \vspace{-4mm}
\end{figure}

\subsection{Performance Regime Structure}
\label{sec:basins}

On the EuRoC and TUM-VI sequences tested, varying $\alpha$ over the candidate set $\mathcal{A} = \{1.0, 2.0, 5.0, 10.0, 20.0\}$ produces outputs that cluster into two groups.
\begin{itemize}
  \item \textsc{imu-trust} ($\alpha \in \{1.0, 2.0\}$). The visual weight is at most twice the calibrated baseline, so the IMU constraint retains greater relative influence. The two values produce similar trajectories on most evaluated sequences.
  \item \textsc{vision-trust} ($\alpha \in \{10.0, 20.0\}$). The visual reprojection cost receives a higher relative weight. The two values produce similar trajectories within this group and differ from the \textsc{imu-trust} group on many evaluated sequences.
    $\alpha\!=\!5.0$ falls in the transition region between the two regimes and is treated separately (see Sec.~\ref{sec:oracle}).
\end{itemize}

The observed grouping is consistent with the change in relative influence between the visual and inertial terms.
In the \textsc{imu-trust} range, the IMU constraint has greater relative influence, and varying $\alpha$ within $\{1.0, 2.0\}$ produces similar converged trajectories.
In the \textsc{vision-trust} range, the visual cost has greater relative influence, and varying $\alpha$ within $\{10.0, 20.0\}$ also produces similar trajectories.
The tested intermediate candidate $\alpha\!=\!5.0$ exhibits transition behavior and is treated separately from the two groups.

On EuRoC MH\_01, both $\alpha \in \{1.0, 2.0\}$ yield $0.138\,\text{m}$ APE, while both values in $\{10.0, 20.0\}$ yield approximately $0.073\,\text{m}$. This corresponds to a $47\%$ difference between the two regimes and a small difference within each regime.
Similar grouping is observed on most evaluated sequences, which motivates the binary prediction used in this work.
Fig.~\ref{fig:basin} illustrates this structure conceptually.

For the output $\alpha$ at inference, we use $\alpha\!=\!2.0$ for \textsc{imu-trust} and $\alpha\!=\!10.0$ for \textsc{vision-trust}.
These values are selected as representative outputs and are the candidates nearest the transition region within their respective groups.

\subsection{Sequential Greedy Label Generation}
\label{sec:oracle}

We generate per-segment binary labels using a sequential non-causal greedy reference based on final full-trajectory APE.
For each training sequence,
\begin{enumerate}
  \item Divide into $S\!=\!10$ equal-length segments by camera frame count.
  \item Initialize the weight of every segment to the released default $\alpha\!=\!1.0$.
  \item For segment $k$, retain the values already selected for segments $1,\ldots,k-1$, evaluate every candidate $\alpha\in\mathcal{A}$ for segment $k$ over $N\!=\!3$ independent tracker runs, and keep future segments $k+1,\ldots,S$ at $\alpha\!=\!1.0$.
    Record the mean final full-trajectory APE for each candidate.
    Averaging over three runs reduces the effect of CUDA non-determinism when candidates produce similar APE values.
  \item Fix the candidate $\alpha_k^{*}$ with the lowest mean final APE and continue to segment $k\!+\!1$.
    Map the selected value to a binary label. The label is \textsc{imu-trust} if $\alpha_k^{*}\in\{1.0,2.0\}$ and \textsc{vision-trust} if $\alpha_k^{*}\in\{10.0,20.0\}$.
    Segments where $\alpha_k^{*}\!=\!5.0$ (the transition region) are excluded from the training set as uncertain.
\end{enumerate}

This forward sequential procedure accounts for interactions with previously selected segments and for the downstream effect of the current choice under the default future schedule.
In a deterministic evaluation, retaining $\alpha\!=\!1.0$ in the candidate set allows each step to preserve the current partial schedule.
Averaging over three CUDA runs reduces, but does not eliminate, small deviations caused by non-determinism.
It is nevertheless not guaranteed to find the globally optimal assignment.
Exhaustive joint optimization would require evaluating all $|\mathcal{A}|^S=5^{10}\approx9.8\times10^6$ combinations per sequence.
We therefore use it as a non-causal greedy reference for label generation, rather than as a strict upper bound.

At inference, \textsc{imu-trust} maps to $\alpha\!=\!2.0$ and \textsc{vision-trust} maps to $\alpha\!=\!10.0$.
On EuRoC, among all SBA calls with a definite binary label (approximately $9{,}500$ of $13{,}700$ total), the greedy reference assigns \textsc{imu-trust} to approximately $76\%$ and \textsc{vision-trust} to $24\%$.
A separate $31\%$ of all SBA calls fall in the transition region ($\alpha_k^{*}\!=\!5.0$) and are excluded from training.
At inference, the MLP always outputs one of the two binary classes.
Calls from segments for which $\alpha\!=\!5.0$ would be near-optimal are therefore assigned according to the learned decision boundary, potentially leading to a suboptimal regime selection.

\subsection{Visual Quality Features}
\label{sec:features}

At each SBA call we assemble a 10-dimensional feature vector $\mathbf{x} \in \mathbb{R}^{10}$ from statistics already available in the tracker as
\begin{equation}
  \mathbf{x} = [
    \underbrace{\ln n_{\mathrm{obs}},\; \ln n_{\mathrm{lm}}}_{\text{quantity}},\;
    \underbrace{\tilde{\sigma}_{u},\; \tilde{\sigma}_{v},\; \ln \Delta t_{\mathrm{imu}}}_{\text{coverage \& timing}},\;
    \underbrace{\ln \bar{d},\; \mathrm{CV}_{d}}_{\text{geometry}},\;
    \underbrace{r_{\mathrm{surv}},\; \ln \bar{e},\; \ln s_{e}}_{\text{stability}}
  ]^{\!\top}
  \label{eq:features}
\end{equation}
All features are drawn from quantities already maintained by the tracker, adding no extra sensing.
The added computation consists of one MLP forward pass per SBA call.
All features reflect tracker state accumulated prior to the current SBA call. Outputs from the current SBA call are not included in the feature vector.
Observation and landmark counts ($\ln n_{\mathrm{obs}}$, $\ln n_{\mathrm{lm}}$) describe the number of constraints contributing to the visual cost.
Pixel scatter ($\tilde{\sigma}_{u}$, $\tilde{\sigma}_{v}$) describes the image-space distribution of features. The preintegration interval ($\ln\Delta t_{\mathrm{imu}}$) relates to the duration over which IMU noise accumulates.
Mean depth ($\ln\bar{d}$) and its spread ($\mathrm{CV}_{d}$) affect the scale and conditioning of reprojection errors relative to the IMU term.
Survival rate ($r_{\mathrm{surv}}$) and reprojection residual statistics ($\ln\bar{e}$, $\ln s_{e}$) provide indicators of tracking quality. Low survival or elevated residuals may be associated with the \textsc{imu-trust} regime.
Log transforms are applied for numerical stability.

\subsection{Visual Quality Estimator}
\label{sec:mlp}

We train a feedforward network $f_{\theta}\colon \mathbb{R}^{10} \to \Delta^{1}$ (probability simplex over two classes) with architecture $10 \to 32 \to 16 \to 2$ (ReLU activations, softmax output, 914 trainable parameters including biases).

\paragraph{Training}
The greedy procedure produces 110 segment-level alpha assignments across the 11 EuRoC sequences (10 per sequence), of which 103 receive a definite binary label (9 \textsc{imu-trust}, 94 \textsc{vision-trust}) and 7 fall in the transition region ($\alpha^*\!=\!5.0$) and are excluded.
The larger proportion of \textsc{vision-trust} labels at the segment level ($91\%$ of labeled segments) is consistent with the visual conditions represented in the EuRoC sequences.
The same greedy procedure applied to TUM-VI Room gives a more balanced distribution. Approximately $57\%$ of labeled segments are \textsc{imu-trust} and $43\%$ are \textsc{vision-trust}.
The difference between the two datasets supports the use of a classifier rather than a single fixed regime.
Assigning each EuRoC segment label to all SBA calls within that segment yields approximately $13{,}700$ feature vectors in total, of which approximately $9{,}500$ have a definite binary label (the remaining $\approx\!31\%$ fall in the transition region and are excluded from training).
The segment-level and call-level class proportions differ because each segment label is repeated according to the number of SBA calls recorded within that segment.
Calls within the same segment share a label and are therefore not independent. The effective label count for EuRoC is consequently 103.
Features are collected from a common baseline ($\alpha\!=\!1.0$) rollout to provide a consistent input distribution across training sequences.
Note that at inference the trajectory may differ from the baseline rollout due to prior MLP decisions, which is a known limitation of this approach.
Training minimizes focal loss~\cite{c_focal} with $\gamma\!=\!2.0$ as
\begin{equation}
  \mathcal{L} = -\sum_{k} w_{y_k}(1 - p_{y_k})^{\gamma} \log p_{y_k}
\end{equation}
Per-class weights are set to inverse class frequency.
Because \textsc{imu-trust} is the majority class ($76\%$ of labeled calls), inverse frequency assigns it a lower base weight than \textsc{vision-trust}.
Before normalization, the \textsc{imu-trust} weight is additionally multiplied by $2$. This adjustment increases its value relative to the unmodified inverse-frequency weight and reflects a design preference against predicting \textsc{vision-trust} under poor visual conditions.
After normalization the final weights are $w_{\mathrm{imu}}\!\approx\!0.78$ and $w_{\mathrm{vis}}\!\approx\!1.22$.
The model is optimized with Adam (learning rate $1\!\times\!10^{-3}$, weight decay $1\!\times\!10^{-4}$, batch size $64$) for $500$ epochs with cosine annealing.
Features are standardized with training-set mean and variance before both training and inference.

\paragraph{Inference}
$f_{\theta}$ runs once per SBA call immediately before the optimization step, consuming only the 10-dimensional feature vector assembled from tracker state prior to the current call.
The predicted class determines $\alpha \in \{2.0,\, 10.0\}$ for that call.
Each SBA call is classified independently using the per-call feature vector.
The segment-level supervision used during training means that calls within the same oracle segment share a label, while at inference the prediction can vary freely across calls.
No temporal smoothing is applied.

\section{EXPERIMENTS}

\begin{table}[t]
  \centering
  \caption{EuRoC MAV RMSE APE [m] (mean over $N\!=\!3$ runs).
    Greedy Ref.\ is the sequential non-causal $S\!=\!10$ reference.
    Smart Fusion uses LOOCV (11-fold).
    $\Delta$ denotes relative change over $\alpha\!=\!1.0$, computed from unrounded means.}
  \label{tab:euroc}
  \setlength{\tabcolsep}{4pt}
  \begin{tabular}{lrrrc}
    \toprule
    Sequence & $\alpha{=}1.0$ & Greedy Ref. & Smart Fusion & $\Delta$ [\%] \\
    \midrule
    MH\_01 & 0.138 & 0.049 & 0.073 & $-47.0$ \\
    MH\_02 & 0.113 & 0.064 & 0.078 & $-30.6$ \\
    MH\_03 & 0.162 & 0.078 & 0.105 & $-35.4$ \\
    MH\_04 & 0.191 & 0.150 & 0.160 & $-16.4$ \\
    MH\_05 & 0.239 & 0.131 & 0.144 & $-40.0$ \\
    V1\_01  & 0.090 & 0.090 & 0.090 & $\phantom{-}0.0$ \\
    V1\_02  & 0.167 & 0.089 & 0.099 & $-40.6$ \\
    V1\_03  & 0.237 & 0.114 & 0.119 & $-49.7$ \\
    V2\_01  & 0.157 & 0.128 & 0.143 & $-8.9$  \\
    V2\_02  & 0.145 & 0.106 & 0.115 & $-20.3$ \\
    V2\_03  & 0.585 & 0.331 & 0.481 & $-17.7$ \\
    \midrule
    \textbf{Mean} & \textbf{0.202} & \textbf{0.121} & \textbf{0.146}
      & $\mathbf{-27.7}$ \\
    \bottomrule
  \end{tabular}
\end{table}

\subsection{Setup}
\paragraph{Datasets}
EuRoC MAV~\cite{c_euroc} contains 11 stereo-inertial sequences in two indoor environments (Machine Hall MH\_01--05 and Vicon Room V1\_01--03, V2\_01--03).
We evaluate on all 11 sequences.
V1\_01 is included but its greedy-reference APE change is 0\%, as varying $\alpha$ over $\mathcal{A}$ produces the same APE ($0.090\,\text{m}$) regardless of which value is selected.
TUM-VI Room~\cite{c_tumvi} provides five fisheye stereo-inertial room sequences (room1--5). We evaluate them under LOOCV to measure performance on a held-out sequence within the dataset.
Room6 is excluded as its ground truth coverage does not span the full trajectory in our setup.

\paragraph{Metric}
RMSE APE [m] with Sim(3) Umeyama alignment~\cite{c_umeyama}.
The $\Delta$ column in all tables is the per-sequence relative change over the $\alpha\!=\!1.0$ baseline, computed from unrounded means.
The Mean row reports arithmetic mean APE across sequences, and its $\Delta$ value is the relative change between the baseline and Smart Fusion mean APEs.
All results are averaged over $N\!=\!3$ independent runs to account for CUDA non-determinism in cuVSLAM.

\paragraph{Baseline and greedy reference}
The primary baseline is the released cuVSLAM configuration with fixed $\alpha\!=\!1.0$.
This comparison measures the change obtained by Smart Fusion relative to the standard fixed-weight configuration.
The sequential greedy procedure in Sec.~\ref{sec:oracle} is reported as a non-causal reference for the candidate set $\mathcal{A}$, not as a globally optimal upper bound.
Each alpha candidate in the greedy search is evaluated over $N\!=\!3$ independent tracker runs, and the mean final APE is used for selection.

\paragraph{Evaluation protocol}
For EuRoC, we use LOOCV over 11 sequences.
In each fold, the MLP is trained on greedy-reference labels from 10 sequences and evaluated on the held-out sequence.
For TUM-VI, we use LOOCV over 5 sequences.
In each fold, the MLP is trained on greedy-reference labels generated from 4 TUM-VI sequences and evaluated on the held-out sequence, using that fold's training statistics for standardization.

\subsection{EuRoC MAV}

Table~\ref{tab:euroc} reports results for all 11 sequences.
Mean APE decreases from $0.202\,\text{m}$ to $0.146\,\text{m}$ ($-27.7\%$), against the greedy-reference result of $0.121\,\text{m}$ ($-40.2\%$).
V1\_01 shows no improvement in either direction, as its APE is identical across all $\alpha$ values.

The largest per-sequence reductions are observed on V1\_03 ($-49.7\%$), MH\_01 ($-47.0\%$), V1\_02 ($-40.6\%$), and MH\_05 ($-40.0\%$).
The reduction achieved by Smart Fusion corresponds to $69\%$ of the difference between the baseline and the greedy reference.
The remaining difference may arise from two sources. The per-SBA feature vector contains only local statistics, while the reference uses full-trajectory APE. In addition, some visual conditions in a held-out fold may be underrepresented in the corresponding training set.
V2\_03 is a notable case where the gap between Smart Fusion ($0.481\,\text{m}$) and the greedy reference ($0.331\,\text{m}$) is largest in absolute terms.
V2\_03 has the highest baseline APE among the evaluated EuRoC sequences. No sequence in the corresponding training fold has a comparable baseline APE.

\subsection{TUM-VI Room}

Table~\ref{tab:tumvi} shows results on TUM-VI Room under LOOCV (5 sequences).
Mean APE decreases from $0.270\,\text{m}$ to $0.230\,\text{m}$ ($-14.9\%$), while the greedy-reference result is $0.222\,\text{m}$ ($-18.0\%$).

Room1 and room3 improve by $29.4\%$ and $21.3\%$, respectively.
Room4 and room5 change by less than $1\%$. The corresponding greedy-reference changes are $6.3\%$ and $0.5\%$, indicating limited improvement over the fixed baseline on these sequences.
The results on held-out TUM-VI sequences suggest that the tracker-internal feature set transfers across sequences within TUM-VI Room without dataset-specific feature definitions.

\begin{table}[t]
  \centering
  \caption{TUM-VI Room RMSE APE [m] (mean over $N\!=\!3$ runs).
    Smart Fusion uses LOOCV (5-fold).
    $\Delta$ denotes relative change over $\alpha\!=\!1.0$, computed from unrounded means.}
  \label{tab:tumvi}
  \setlength{\tabcolsep}{4pt}
  \begin{tabular}{lrrrc}
    \toprule
    Sequence & $\alpha{=}1.0$ & Greedy Ref. & Smart Fusion & $\Delta$ [\%] \\
    \midrule
    room1 & 0.286 & 0.194 & 0.202 & $-29.4$ \\
    room2 & 0.148 & 0.138 & 0.143 & $-3.2$  \\
    room3 & 0.525 & 0.391 & 0.413 & $-21.3$ \\
    room4 & 0.104 & 0.098 & 0.103 & $-0.7$  \\
    room5 & 0.289 & 0.287 & 0.289 & $\phantom{-}0.0$  \\
    \midrule
    \textbf{Mean} & \textbf{0.270} & \textbf{0.222} & \textbf{0.230}
      & $\mathbf{-14.9}$ \\
    \bottomrule
  \end{tabular}
\end{table}

\section{CONCLUSIONS}

Smart Fusion formulates per-SBA $\alpha$ selection as binary classification based on two empirical performance regimes observed over the tested candidate set.
A 914-parameter MLP trained on sequential greedy labels from tracker-internal features reduces mean RMSE APE by $27.7\%$ on EuRoC (LOOCV, 11 sequences) and $14.9\%$ on TUM-VI Room (LOOCV, 5 sequences). These reductions correspond to $69\%$ and $83\%$ of the respective differences between the baseline and the greedy reference.
The remaining differences may be related to the mismatch between the local per-SBA feature vector and the full-trajectory APE objective, as well as limited training coverage of difficult operating conditions. Future work could examine temporal features and a broader range of training conditions.

\section*{ACKNOWLEDGMENT}
The authors thank the cuVSLAM team at NVIDIA for making the system available as open source.


\end{document}